%% file: index.tex
\documentclass[11pt]{article}

\usepackage[final]{acl}

\usepackage{times}
\usepackage{latexsym}

\usepackage[T1]{fontenc}

\usepackage[utf8]{inputenc}

\usepackage{microtype}

\usepackage{inconsolata}

\usepackage{graphicx}

\input{preamble_common}
\input{dataset_stats}

\title{Critic's Eye: Review-Grounded Graph Reasoning for Long-Form Motif Discovery}

\author{
	\textbf{Xinyu Sun\textsuperscript{1,2}},
	\textbf{Zhangyue Yin\textsuperscript{1}},
	\textbf{Xinyi Ju\textsuperscript{1,2}},
	\textbf{Xin Wang\textsuperscript{1,2}}
	\\
	\\
	\textsuperscript{1}College of Computer Science and Artificial Intelligence\\ Fudan University, \\
	\textsuperscript{2}Shanghai Key Laboratory of Intelligent Information Processing
	\\
	\small{
		\href{mailto:sunxy23@m.fudan.edu.cn}{sunxy23@m.fudan.edu.cn},
		\href{mailto:yinzy21@m.fudan.edu.cn}{yinzy21@m.fudan.edu.cn},
		\href{mailto:juxinyi.janice@gmail.com}{juxinyi.janice@gmail.com},
		\href{mailto:xinw@fudan.edu.cn}{xinw@fudan.edu.cn}
	}
}

\hypersetup{
	pdftitle={Critic's Eye: Review-Grounded Graph Reasoning for Long-Form Motif Discovery},
	pdfauthor={Xinyu Sun, Zhangyue Yin, Xinyi Ju, Xin Wang}
}

\begin{document}

	\maketitle
	\begin{abstract}
		\input{abstract}
	\end{abstract}

	\section{Introduction}

	\input{intro}

	\section{Related Work}

	\input{related_work}

	\section{Method}
	\input{method}

	\begin{table}[t]
		\centering
		\small
		\vspace{-0.8em}

		\vspace{-0.8em}
		\begin{tabular}{l r | l r}
		\toprule
		\textbf{Node Type} & \textbf{Count} & \textbf{Edge Type} & \textbf{Count} \\
		\midrule
		Act / Scene & \dataNumActs{} / \dataNumScenes{} & Hierarchical ($\mathcal{E}_{\textsc{hier}}$) & \dataEdgesHier{} \\
		Speech Section & \dataNumSections{} & Sequential ($\mathcal{E}_{\textsc{seq}}$) & \dataEdgesSeq{} \\
		Atomic Line & \dataTotalLines{} & Metadata ($\mathcal{E}_{\textsc{agent}}$) & \dataEdgesAgent{} \\
		Character & \dataNumChars{} & \textbf{Total Explicit} & \textbf{\dataEdgesExplicit{}} \\
		\midrule
		\textit{Latent Motif} & \dataNumMotifs{} & \textit{Thematic Link} & \dataNumThematicLinks{} \\
		\bottomrule
		\end{tabular}
		\caption{Statistics of the \textit{Macbeth} Narrative Graph.
		\textit{Thematic Link} counts every criticism-derived edge in the graph,
		including relations between two textual units and relations anchored to a speech or scene rather than to a single line.
		The line$\rightarrow$motif membership links that supply supervision and evaluation gold are the subset plotted in Figure~\ref{fig:longtail}
		(\dataNumAnnotations{} links over \dataNumQueryLines{} lines and \dataNumActiveMotifs{} motifs).}
		\label{tab:graph_stats}
	\end{table}

	\begin{table}[t]
		\centering
		\vspace{-0.8em}
		\small
		\setlength{\tabcolsep}{4pt}
		\renewcommand{\arraystretch}{1.35}
		\begin{tabularx}{\columnwidth}{@{}l X@{}}
		\toprule
		\textbf{Stratum (Type)} & \textbf{Focus \& Role in Narrative Graph} \\
		\midrule

		\textbf{1. Humanist} & Focus: Character agency, ethical conflicts, and tragedy. \\
		\textit{(\dataCorpusHumanist{} Monographs} & \textit{Role}: \textbf{High-Density Supervision.} Comprising foundational book-length studies and major critical volumes. \\
		\textit{\& Volumes)} & \textit{Key Sources:} \citet{bradley1904shakespearean}, \citet{bloom1987william}, \citet{fletcher1844macbeth}, \citet{hazlitt1903macbeth}, \citet{weiss1876wit}. \\

		\textbf{2. Socio-Cultural} & Focus: Folklore, politics, history, and theatre practice. \\
		\textit{(\dataCorpusSocio{} Mixed Sources)} & \textit{Role}: \textbf{Contextual Anchoring.} Captures world-building motifs (e.g., \textit{Witchcraft}, \textit{Kingship}) missed by pure text analysis. \\
		\textit{(Books/Essays)} & \textit{Key Sources:} \citet{dyer1883folklore} (Folklore), \citet{adams1917playhouses} (Theatre), \citet{tolstoy1906shakespeare} (Counter-criticism), \citet{moschovakis2008macbeth}. \\

		\textbf{3. Phenomenological} & Focus: Sensory imagery, objects, and embodied cognition. \\
		\textit{(\dataCorpusPhenom{} Articles)} & \textit{Role}: \textbf{Latent Discovery.} Forces the model to recover non-explicit associations (e.g., \textit{Smell}, \textit{Dread}). \\
		\textit{(Specific Studies)} & \textit{Key Sources:} \citet{harris2007smell}, \citet{spolsky2011embodied}, \citet{cheung1984dread}, \citet{sachon2020shakespeare}. \\

		\bottomrule
		\end{tabularx}
		\caption{\textbf{Stratification of the Critical Corpus.}
		Sources are categorized by interpretive lens.
		Note that the \textit{Humanist} stratum comprises of extensive monographs and volumes,
		providing a dense backbone of character analysis, while other strata introduce targeted, fine-grained perspectives.
		Full bibliographic details are in Appendix~\ref{app:corpus}.
		}
		\label{tab:corpus_stratification}
		\vspace{-0.8em}
	\end{table}

	\section{The \textsc{MotifBench} Dataset}
	\label{sec:dataset}
	\begin{table*}[!t]
		\centering
		\small
		\setlength{\tabcolsep}{2.5pt}
		\renewcommand{\arraystretch}{1.05}
		\vspace{-0.8em}

		\begin{tabularx}{\textwidth}{@{}l *{4}{>{\centering\arraybackslash}X} *{4}{>{\centering\arraybackslash}X} >{\centering\arraybackslash}X@{}}
		\toprule
		\textbf{Method}
		& \multicolumn{4}{c}{\textbf{Hit}}
		& \multicolumn{4}{c}{\textbf{nDCG}}
		& \textbf{MRR}\\
		\cmidrule(lr){2-5}\cmidrule(lr){6-9}
		& @1 & @3 & @5$^{*}$ & @10
		& @1 & @3 & @5 & @10
		& \\

		\midrule
		\multicolumn{10}{l}{\textit{\textbf{Embedding SOTA} (Best of dense retrievers)}} \\
		Sentence-BERT
		& 0.042 & 0.074 & 0.106 & 0.106
		& 0.041 & 0.053 & 0.064 & 0.074
		& 0.081 \\

		\midrule
		\multicolumn{10}{l}{\textit{\textbf{Open-Source LLMs} (Highlighting Inverse Scaling)}} \\
		GPT-OSS-20B (+RAG)
		& 0.042 & 0.088 & 0.117 & 0.120
		& 0.042 & 0.070 & 0.081 & 0.082
		& 0.070 \\
		GPT-OSS-120B (+RAG) $^{\dagger}$
		& 0.032 & 0.064 & 0.085 & 0.085
		& 0.032 & 0.049 & 0.058 & 0.058
		& 0.050 \\
		DeepSeek-R1-70B (Zero-shot)
		& 0.081 & 0.124 & 0.134 & 0.159
		& 0.081 & 0.106 & 0.110 & 0.118
		& 0.105 \\

		\midrule
		\multicolumn{10}{l}{\textit{\textbf{Proprietary Foundation Models} (Highlighting Semantic Ceiling)}} \\
		GPT-5.2-Chat (Zero-shot)
		& 0.064 & 0.127 & 0.170 & 0.184
		& 0.064 & 0.099 & 0.116 & 0.120
		& 0.100 \\
		Claude-Opus-4.5 (Zero-shot)
		& 0.078 & 0.148 & 0.198 & 0.276
		& 0.078 & 0.118 & 0.138 & 0.163
		& 0.129 \\

		Gemini-3-pro-preview (+RAG)
		& \underline{0.088} & \underline{0.156} & \underline{0.226} & \underline{0.332}
		& \underline{0.088} & \underline{0.129} & \underline{0.156} & \underline{0.191}
		& \underline{0.148} \\

		\midrule
		\textbf{\ce (Ours)}
		& \textbf{0.548} & \textbf{0.774} & \textbf{0.848} & \textbf{0.922}
		& \textbf{0.548} & \textbf{0.646} & \textbf{0.682} & \textbf{0.716}
		& \textbf{0.676} \\

		\bottomrule
		\end{tabularx}

		\begin{minipage}{\textwidth}
			\vspace{1mm}
			\footnotesize
			\textit{Note:} We report the \textbf{top-performing model} for each category.
			$^{\dagger}$ \textbf{GPT-OSS-120B} is included to illustrate the \textit{inverse scaling} phenomenon (underperforming the 20B model due to instruction drift).
			Underlines denote the best baseline performance. $^{*}$ indicates the main metric we focus on (Hit@5).
			Full results for all 12 evaluated models are provided in Appendix ~\ref{app:full_results}.
		\end{minipage}
		\vspace{-0.8em}

		\caption{\textbf{Performance Comparison with Frontier Models.}
		Even the strongest proprietary model (Gemini-3-pro-preview) hits a \textbf{ceiling} ($\sim$\resPropCeilingHitFive{} Hit@5),
		lagging significantly behind \ce's structural reasoning.}
		\label{tab:main_results}
		\vspace{-0.8em}

	\end{table*}

	\input{dataset}

	\section{Experiments}
	\input{experiment}

	\section{Conclusion}
	\input{conclusion}

	\section*{Limitations}
	\input{limit}

	\section*{Ethics Statement}
	\input{ethics}

	\section*{Data and Code Availability}
	\input{availability}

	\section*{Acknowledgments}
	We gratefully acknowledge the Folger Shakespeare Library for providing the high-quality TEI-encoded digital texts that serve as the narrative backbone of this study.
	We also thank Project Gutenberg and its volunteers for their enduring commitment to preserving the public domain criticism used in our dataset.

	\bibliography{refs}

	\appendix

	\section{Detailed List of the Critical Corpus}
	\label{app:corpus}
	\input{appendix/corpus}

	\section{Motif Extraction Prompts}
	\label{app:prompts}
	\input{appendix/prompts}

	\section{Reproducibility Details}
	\label{app:reproducibility}
	\input{appendix/reproducibility}

	\section{Full Benchmark Results}
	\label{app:full_results}
	\input{appendix/fullresult}

	\section{Qualitative Case Studies}
	\label{app:case_studies}
	We provide the full textual context and model outputs for the case studies discussed in Section 5.4.
	\input{appendix/casestudy}

\end{document}

%% file: preamble_common.tex
\newcommand{\ce}{\textsc{Critic's Eye}\xspace}

\newcommand{\sjuzet}{\textbf{\textit{sju\v{z}et}}\xspace}
\newcommand{\fabula}{\textbf{\textit{fabula}}\xspace}

\newcommand{\ssjuzet}{\textit{sju\v{z}et}\xspace}
\newcommand{\ffabula}{\textit{fabula}\xspace}

\newcommand{\eg}[1]{\textit{e.g. #1}}

\usepackage{xspace}
\usepackage{standalone}
\usepackage{siunitx}
\usepackage{float}
\usepackage{graphicx}
\usepackage{caption}
\usepackage{dblfloatfix}
\usepackage[utf8]{inputenc}
\usepackage[T1]{fontenc}
\usepackage{mathtools}
\usepackage{amsmath}
\usepackage{amssymb}
\usepackage{booktabs}
\usepackage{array}
\usepackage{tabularx}
\usepackage{ragged2e}
\usepackage{enumitem}
\usepackage{makecell}
\usepackage{threeparttablex}
\usepackage{tcolorbox}

%% file: dataset_stats.tex
\newcommand{\dataNumMotifs}{281}
\newcommand{\dataTotalLines}{2,456}
\newcommand{\dataNumSections}{865}
\newcommand{\dataNumChars}{44}
\newcommand{\dataNumActs}{5}
\newcommand{\dataNumScenes}{28}
\newcommand{\dataEdgesHier}{3,397}
\newcommand{\dataEdgesSeq}{2,602}
\newcommand{\dataEdgesAgent}{1,058}
\newcommand{\dataEdgesExplicit}{7,057}
\newcommand{\dataNumThematicLinks}{923}

\newcommand{\dataNumAnnotations}{351}
\newcommand{\dataNumQueryLines}{283}
\newcommand{\dataNumActiveMotifs}{193}

\newcommand{\dataTailCount}{187}
\newcommand{\dataTailPct}{96.9\%}
\newcommand{\dataSingletons}{140}
\newcommand{\dataSingletonPct}{73\%}
\newcommand{\dataImbalanceRatio}{35.0}

\newcommand{\dataNumDocs}{26}
\newcommand{\dataCorpusHumanist}{12}
\newcommand{\dataCorpusSocio}{8}
\newcommand{\dataCorpusPhenom}{6}

\newcommand{\resCEHitFive}{84.8\%}
\newcommand{\resGeminiHitFive}{22.6\%}
\newcommand{\resPropCeilingHitFive}{23\%}
\newcommand{\resOpenCeilingHitFive}{13\%}

\newcommand{\resAblRandomHitFive}{0.31}
\newcommand{\resAblHeteroHitFive}{0.59}
\newcommand{\resAblSeqHitFive}{0.76}
\newcommand{\resAblFullHitFive}{0.85}
\newcommand{\resAblPretrainGain}{+0.53}

%% file: abstract.tex
Deep narrative understanding requires distinguishing \textit{what} happens from \textit{how} it is presented,
moving beyond the chronological \ffabula to decode the \ssjuzet,
the discursive organization through which themes are staged.
We study this through the recurrence of \emph{literary} motifs (\textit{blood} in \textit{Macbeth}, for instance),
where symbolic significance emerges not from local cues, but through repetition-with-variation across the whole work.
We operationalize this setting as Computational Motif Discovery,
a transductive task that predicts missing \textit{line}$\rightarrow$\textit{motif} links using the full narrative structure of a play.
We propose \ce, which models the work as a heterogeneous Narrative Topology Graph and performs discriminative inference over global structural evidence.
\ce achieves \resCEHitFive{} Hit@5 on our benchmark, nearly quadrupling state-of-the-art proprietary foundation models (Gemini-3-pro-preview, $\sim$\resGeminiHitFive{}).
Our analysis reveals that despite massive scaling, sequence-based models hit a ``semantic ceiling'', struggling to resolve dispersed dependencies.
These findings suggest that explicit structural priors are a far more effective inductive bias than parameter scaling for decoding the complex architecture of long-form literature.

%% file: intro.tex
\begin{figure}[!t]
    \vspace{-0.8em}
    \centering
    \includegraphics[width=\linewidth,  scale=0.5]{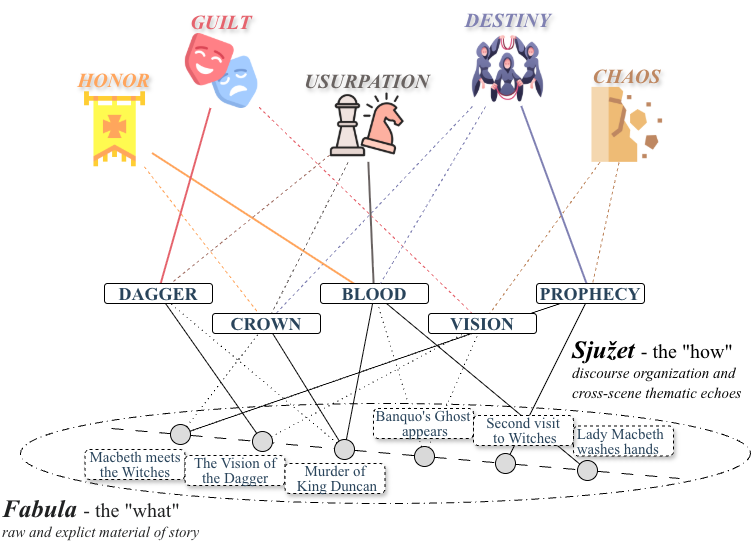}
    \vspace{-0.8em}
    \caption{\textbf{Why motif evidence is spread out.}
    \footnotesize
    Bottom: the \fabula, the play's events in chronological order.
    Middle: recurring images, each attaching to events that lie far apart on that order.
    Top: the themes those images carry.}
    \label{fig:gap}
    \vspace{-1em}
\end{figure}

A fundamental challenge in narrative intelligence is distinguishing the raw material of a story from its artistic presentation.
This distinction is formalized by Russian Formalism as the duality between the \fabula (the chronological sequence of events)
and the \sjuzet (the discursive organization of the text) \cite{tomashevsky1965thematics, propp1968morphology, bal2009narratology}.
While the \ffabula follows the linear causality of ``what happened,''
the \ssjuzet encodes the topological logic of ``how it is presented,''
employing techniques such as recurrence, focalization, and anachrony to stage meaning.

A particularly central form of \ssjuzet-level evidence is \textbf{motif recurrence}.
A motif is an element the work keeps returning to, in varied form, and its instances are read together:
each carries the weight it does because of the others.
\textit{Macbeth} has \textit{blood} (Figure~\ref{fig:gap}), the murder of sleep, and the washing of hands.
The \ffabula and the \ssjuzet are unequally sensitive to the order of presentation.
Presenting the \ffabula in a different order alters what a reader knows and when, and with it the rhetorical effect of the telling
\cite{genette1980narrative, sternberg1978expositional},
without altering the thematic content of the work \cite{tomashevsky1965thematics}.
A motif therefore cannot be individuated by its position in the telling, nor by surface form, since its instances need share no wording,
which is why \citet{spurgeon1935shakespeares} assembles an image across a whole play rather than at any single point in it.
Meaning at the \ssjuzet level is always distributed and relational in this sense,
emerging only from how elements recur and interact across a work.

This conception aligns with a longstanding aspiration in NLP:
to move beyond surface processing toward models that can resolve ambiguity and construct coherent meaning.
However, dominant task formulations have largely emphasized \ffabula-level inference, favoring explicit,
locally supervised content such as event extraction \cite{sun2024event}, plot summarization, or attribute summarization \cite{yuan2024evaluating}.
While effective for factual reporting,
such objectives inherently flatten the rich, non-linear semantics of literature.
By stripping away the discursive organization,
current models neglect the \ssjuzet, missing the critical interpretative cues required to decode a work's deeper intent.

We operationalize this missing component through \textbf{Computational Motif Discovery}.
Unlike explicit events that follow linear time, motifs are structural anchors determined by their distribution across the whole work:
they induce long-range bindings that connect distant moments into a thematic configuration.
As \citet{spurgeon1935shakespeares} observes,
motifs consist of ideas that are "interwoven," forming a structure "subtle and complex" that defies simple plot mechanics.
Their meanings are often non-stationary and context-dependent (e.g., \texttt{blood} shifting from \textit{honor} to \textit{guilt} across acts),
making motif resolution poorly matched to surface-level retrieval or inductive classification.
This makes motif resolution a particularly clean testbed for \ssjuzet-level inference:
the target is not to summarize content, but to recover non-local bindings induced by recurrence-with-variation.

This setting exposes the intrinsic limits of sequence-based modeling.
Despite their surface fluency, Large Language Models (LLMs) fundamentally process text as flat streams.
Recent empirical studies indicate that LLMs struggle with long-horizon coherence \cite{tian2024are}
and planning \cite{senanayake2025language},
often failing to integrate evidence that is distributed across thousands of lines ("lost-in-the-middle") \cite{levy2024same}.
In our experiments, even state-of-the-art proprietary foundation models (e.g., Claude, GPT-series, Gemini) hit a distinct ``ceiling'' on motif resolution.
Despite massive parameter scaling and likely pre-training exposure to the source text, they achieve a Hit@5 of only \textbf{$\sim$\resPropCeilingHitFive{}},
lagging far behind our approach (\textbf{\resCEHitFive{}}).
This suggests that the bottleneck is not computational capacity or local semantics, but the lack of explicit topological priors required to model global recurrence.

To address this, we propose \ce, a framework that models narrative not as a sequence, but as a heterogeneous \textbf{Narrative Topology Graph}.
By explicitly encoding the interplay of lines, characters, and scenes, we formulate motif resolution as "transductive link discovery",
enabling the model to propagate weak thematic signals across the global structure to resolve interpretive ambiguities.

%% file: related_work.tex
\paragraph{Modeling the \textit{Fabula}: Entities and Events.}
A robust literature in NLP models explicit narrative content, the "what happened", through entities, character representations, and event structures.
On the character side, prior work modeled discrete roles and archetypes \cite{bamman2014bayesian, stammbach2022heroes},
while recent efforts expand toward character profiling with LLMs,
summarizing attributes and relationships from broader context \cite{yuan2024evaluating}.
Related directions include persona-driven decision modeling \cite{xu2025character} and goal or intent induction from semantic encodings \cite{rahimtoroghi2017modelling}.
On the event side, causality has been identified as central to story understanding:
\citet{sun2024event} study open-world event causality,
building on earlier paradigms for event chain extraction and temporal dynamics \cite{chambers2009unsupervised, fang2024transformer}.
Recent benchmarks continue to push for narrative-driven understanding in drama series \cite{zhang2025drama}.
Across these lines, supervision and evaluation typically favor locally explicit evidence (e.g., event mentions, pairwise relations, span-level answers),
emphasizing plot-level reconstruction over interpretive meaning.

\begin{figure*}[!t]
    \vspace{-0.8em}
    \centering
    \includegraphics[width=\textwidth]{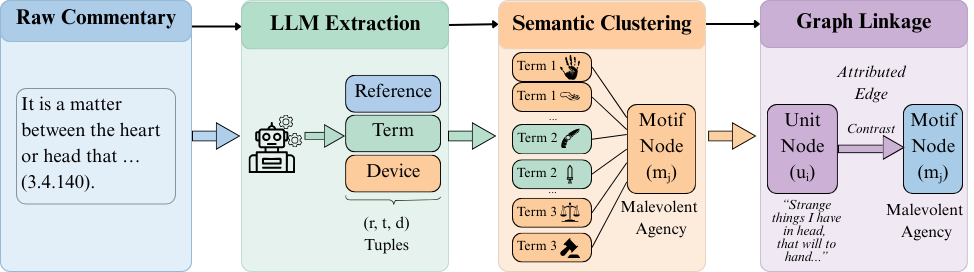}
    \vspace{-0.8em}
    \caption{\textbf{The \ce Graph Construction Pipeline.}
    We fuse the explicit \textbf{Narrative Backbone} (parsed from TEI XML) with a latent \textbf{Thematic Layer} derived from literary criticism.
    (1) An LLM extracts structured tuples (Reference, Theme, Device) from unstructured commentary.
    (2) These mentions are semantically clustered into canonical motif nodes.
    (3) The resulting heterogeneous graph encodes both the linear \fabula and the critical \ssjuzet, serving as the substrate for transductive inference.
    The extraction prompt and output schema are in Appendix~\ref{app:prompts}.
    }
    \label{fig:data_pipeline}
    \vspace{-0.8em}

\end{figure*}

\paragraph{Modeling the \ssjuzet: Discourse and Long-Form Coherence.}
Beyond fabula, other work attempts to capture narrative discourse, how a story is presented, by operationalizing discursive features.
\citet{piper2024using} employ LLMs to annotate deictic dimensions such as time, setting, and perspective,
and \citet{zhao2024narrativeplay} leverage narrative structures to guide interactive role-play.
In generation and evaluation,
several studies analyze the discourse-level properties of LLM stories and identify deficits in narrative planning and arc control \cite{tian2024are, senanayake2025language},
or struggles in decomposing complex narratives into structural scene graphs \cite{yang2025llm}.
Metrics for long-form coherence,
such as BooookScore \cite{chang2024booookscore},
provide useful signals but still face fundamental challenges when models process text as flat sequences under finite context windows,
These include ``lost-in-the-middle'' effects \cite{levy2024same, baker2024lost},
and inverse scaling phenomena on complex reasoning tasks \cite{mckenzie2023inverse}.
More broadly, widely used NLU benchmarks tend to privilege fact-centric objectives (e.g., extractive or multi-hop QA),
rewarding locally explicit answers over global interpretive coherence \cite{rajpurkar2016squad, yang2018hotpotqa}.

\paragraph{From Extraction to Topological Discovery.}
We move beyond detecting isolated elements (characters, events, discursive tags) to model the latent topology that binds them across a whole work.
While prior work has utilized graphs for explicit event relations \cite{fang2024transformer},
temporal forecasting \cite{cai2025segnn},
or logic-constrained completion \cite{huang2025logic},
these methods often rely on relational assumptions (e.g., decay, transitivity) that do not transfer to literary analysis.
We focus on \textbf{Computational Motif Discovery}:
inferring motif associations whose evidence is ``interwoven'' \cite{spurgeon1935shakespeares}
and distributed through recurrence-with-variation rather than localized mentions.
This task targets narrative \emph{linking functions} \cite{herman2003narrative}
and requires resolving non-stationary, structurally mediated dependencies:
a setting naturally cast as whole-work, transductive link discovery rather than local sequence prediction.

%% file: method.tex
\paragraph{Task.}
Let $\mathcal{M}$ be a fixed motif inventory (Section~\ref{subsec:instantiation}).
Given a query unit $u \in \mathcal{U}$, a system scores every $m \in \mathcal{M}$ as $s_{\theta}(u,m)$
and returns the induced ranking of the full inventory
(the inventory is provided at training and inference time).
Performance is measured by the ranks assigned to the motifs criticism attaches to $u$.
We use ``And pity, like a naked newborn babe'' (\textit{Macbeth}~1.7.21) as a running example;
Section~\ref{subsec:case_study} and Appendix~\ref{app:case_studies} return to it.

\subsection{Formalizing the Narrative Space} \label{subsec:formalize}
A literary narrative is not merely a linear sequence of tokens, but a hierarchical system composed of interacting entities.
To capture this, we adopt a formalization grounded in structural narratology \cite{bal2009narratology, genette1980narrative},
mapping the components of narrative discourse to an observable backbone tuple $\mathcal{N} = (\mathcal{U}, \mathcal{A}, \mathcal{S})$:

\begin{itemize}[noitemsep, topsep=0pt]
    \item \textbf{Atomic Units ($\mathcal{U}$): The Micro-Level Voice.}
        Let $\mathcal{U} = \{u_1, ..., u_n\}$ denote atomic textual units (\eg {lines in drama, sentences in novels}),
        ordered by their presentation in the text.
        These correspond to Genette's concept of \textit{voice}, the immediate textual surface where motifs are realized.
        These units are strictly ordered by their linear presentation ($t=1 \dots n$).
    \item \textbf{Agents ($\mathcal{A}$): The Meso-Level Persona.}
        Let $\mathcal{A} = \{a_1, ..., a_k\}$ denote narrative agents (\eg {characters, narrators}).
        Following \citet{bamman2014bayesian}, we model characters not as static attributes but as \textit{bundles of habits} defined by their utterances.
        We define a partial speaker mapping $\pi: \mathcal{U} \to \mathcal{A}\cup\{\bot\}$, connecting each atomic unit to its originating persona,
        with $\bot$ for units that have no single speaker.
    \item \textbf{Structural Containers ($\mathcal{S}$): The Macro-Level Mood.}
    $\mathcal{S}$ represents the spatiotemporal hierarchy (\eg{acts, scenes}).
    Analogous to narrative \textit{mood} or setting, these containers impose boundaries on interaction.
    Formally, they form a rooted tree via a parent function $p:\mathcal{S}\to\mathcal{S}\cup\{\varnothing\}$,
    with an assignment $\rho:\mathcal{U}\to\mathcal{S}$ anchoring each unit to its minimal container.
\end{itemize}

All components in $(\mathcal{U}, \mathcal{A}, \mathcal{S})$ are explicitly observable from the source text,
forming the topological scaffold upon which thematic reasoning operates.

\begin{figure*}[!t]
    \vspace{-0.8em}
    \centering
    \includegraphics[width=\textwidth]{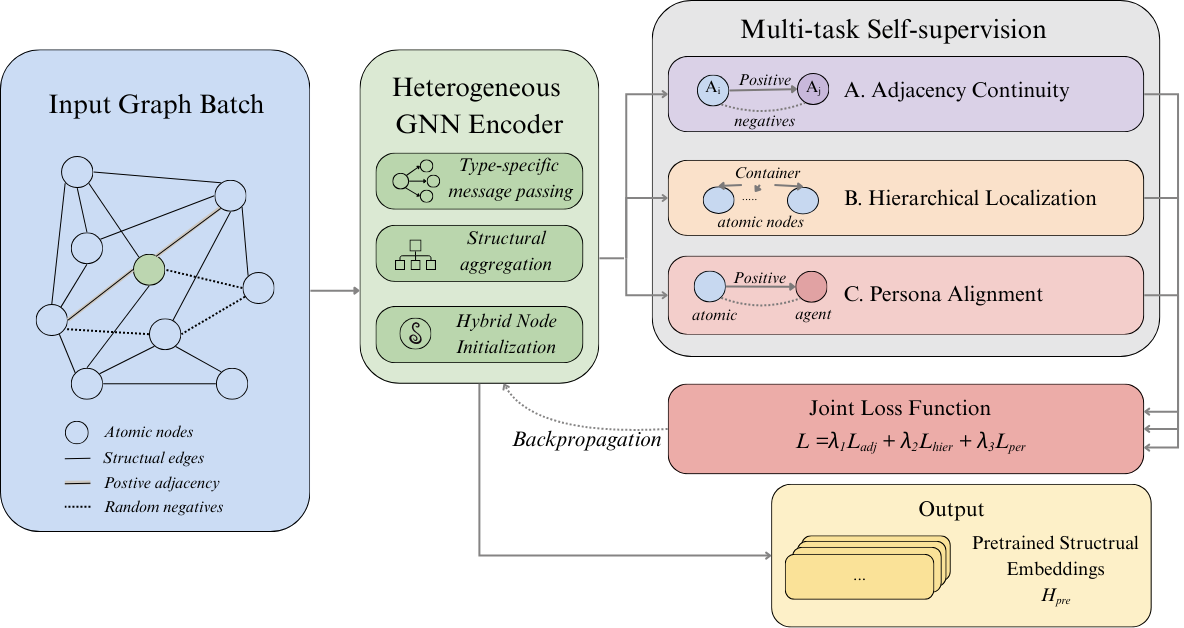}
    \vspace{-0.8em}
    \caption{\textbf{Stage 1: Multi-Task Structural Pre-training.}
    Before accessing motif labels, the encoder is forced to internalize the narrative ``anatomy'' via three intrinsic objectives:
    \textbf{(A) Adjacency Continuity} ($u_t \to u_{t+1}$) captures local discourse flow;
    \textbf{(B) Hierarchical Localization} ($u \to s$) recovers macro-structural positioning;
    and \textbf{(C) Persona Alignment} ($u \to a$) learns character-specific stylistic features.
    This \textit{structure-first} strategy aligns the embedding space with the narrative topology. }
    \label{fig:pretrain}
    \vspace{-0.8em}
\end{figure*}

\subsection{Constructing the Narrative Topology} \label{subsec:topology}
We operationalize $\mathcal{N}$ as a heterogeneous narrative topology graph $\mathcal{G}=(\mathcal{V},\mathcal{E})$,
transforming the theoretical schema into a computable structure.
Let $\mathcal{V}=\mathcal{U}\cup\mathcal{A}\cup\mathcal{S}$.

\begin{enumerate}
    \item \textbf{Textual flow ($\mathcal{E}_{\textsc{seq}}$):}
    $(u_t \xrightarrow{\textsc{seq}} u_{t+1})$ captures local discourse flow in presentation order.
    We initialize node features using a frozen sentence encoder to preserve local semantics.

    \item \textbf{Hierarchical composition ($\mathcal{E}_{\textsc{hier}}$):}

    Edges $(u \xrightarrow{\textsc{hier}} \rho(u))$ and $(s \xrightarrow{\textsc{hier}} p(s))$ encode the vertical organization of the \ssjuzet,
    creating topological shortcuts that allow information to propagate across long temporal distances within the same scene.

    \item \textbf{Agentive interaction ($\mathcal{E}_{\textsc{agent}}$):}
    $(\pi(u) \xrightarrow{\textsc{agent}} u)$ model the character-driven nature of drama, allowing the model to learn persona-specific stylistic features.
\end{enumerate}

\subsection{Motif Instantiation via Critical Clustering} \label{subsec:instantiation}

To instantiate motif nodes and sparse positive links,
we leverage literary criticism as a source of expert supervision.
Figure~\ref{fig:data_pipeline} shows the three steps below as a single pipeline, from criticism to graph.
Since critics describe identical themes using varied terminology (e.g., \textit{blood}, \textit{gore}, \textit{sanguine imagery}),
we employ a \textbf{Schema-Constrained Extraction} pipeline.

\begin{enumerate}
    \item \textbf{Extraction:} A schema-constrained LLM (GPT-OSS-120B, temperature 0) extracts tuples
    $(r, t, d)$ from cited critic prose, where $r$ is a resolvable reference, $t$ is a thematic
    descriptor, and $d$ is a rhetorical device tag (\eg \textit{Metaphor}). Tuples failing schema
    validation or citation resolution are discarded; a manual audit of 200 extracted tuples
    indicates over 90\% semantic precision (prompt and schema in Appendix~\ref{app:prompts}).
    \item \textbf{Clustering:} We cluster descriptors $t$ into canonical motif nodes $\mathcal{M}$ by
    agglomerative clustering over BGE-large-en-v1.5 embeddings (average linkage on cosine distance,
    $\varepsilon = 0.30$).
    \item \textbf{Attributed Linkage:} We instantiate observed edges $\mathcal{E}_{obs}$ between units
    $u$ and motifs $m$. Repeated assertions of the same pair are merged, so $\mathcal{E}_{obs}$ is a set
    of distinct line--motif pairs rather than a multiset of mentions.
\end{enumerate}

$\mathcal{M}$ is therefore a fourth node type of $\mathcal{G}$, alongside $\mathcal{U}$, $\mathcal{A}$ and $\mathcal{S}$,
and the only one not observable in the text.

\subsection{Two-Stage Structure-to-Semantics Transfer} \label{subsec:transfer}
\paragraph{The encoder.}
$f_{\theta}$ is a two-layer, 768-dimensional heterogeneous graph network following \citet{hu2020hgt}.
Each relation type of $\mathcal{E}$ carries its own projection,
and the messages reaching a node over its incident relation types are summed.
Units are initialized from a frozen sentence encoder; agents and containers carry learned embeddings.
Optimization and hardware settings are in Appendix~\ref{app:reproducibility}.

Our training pipeline adopts a Structure-First strategy,
reflecting the hypothesis that literary meaning is constrained by topological position before it is filled with semantic content.

\paragraph{Stage 1: Multi-Task Structural Pre-training.}
We force the encoder to internalize the narrative "anatomy" via three self-supervised objectives (Fig. \ref{fig:pretrain}),
without accessing motif labels:
\begin{itemize} [noitemsep, topsep=0pt]
\item \textbf{Adjacency Continuity ($\mathcal{L}_{seq}$):} cross-entropy over $\mathcal{U}$, predicting $u_{t+1}$.
\item \textbf{Hierarchical Localization ($\mathcal{L}_{hier}$):} cross-entropy over $\mathcal{S}$, predicting $\rho(u)$.
\item \textbf{Persona Alignment ($\mathcal{L}_{agent}$):} cross-entropy over $\mathcal{A}$, predicting $\pi(u)$ where $\pi(u)\neq\bot$.
\end{itemize}
The encoder $f_{\theta}$ minimizes $\mathcal{L}_{pre} = \lambda_{1}\mathcal{L}_{seq} + \lambda_{2}\mathcal{L}_{hier} + \lambda_{3}\mathcal{L}_{agent}$,
mapping nodes to a structure-aware topological space.

\paragraph{Stage 2: Semantic Alignment via Contrastive Learning.}
We then align this topological space with the semantic space of motifs.
Using a frozen Motif Encoder $g_{\phi}$ (e.g., BGE), we treat motif discovery as a dense retrieval task.
We optimize an InfoNCE loss to maximize the similarity between the structure-aware unit embedding $\mathbf{h}_{atomic} = f_{\theta}(u)$ and its ground-truth motif embedding $\mathbf{h}_{motif}$:
\begin{equation}
\mathcal{L}_{\text{align}} = - \log \frac{\exp(s(\mathbf{h}_{\text{atomic}}, \mathbf{h}^{+}_{\text{motif}}) / \tau)}{\sum_{j \in \mathcal{N}} \exp(s(\mathbf{h}_{\text{atomic}}, \mathbf{h}^{j}_{\text{motif}}) / \tau)}
\end{equation}
This effectively "anchors" the floating semantic concepts to the rigid narrative topology.

%% file: dataset.tex
To evaluate \ce, we constructed \textsc{MotifBench}, a specialized graph dataset centered on Shakespeare's \textit{Macbeth}.
Unlike crowd-sourced benchmarks, our ground truth is derived from authoritative literary criticism.
The dataset comprises two distinct layers: the explicit \textbf{Narrative Backbone} and the latent \textbf{Critical Layer}.

\subsection{The Narrative Backbone}
We parse the TEI-encoded XML of \textit{Macbeth} from the Folger Shakespeare Library \cite{folger2025macbeth} into a heterogeneous graph $\mathcal{G}$.

The node schema corresponds to the play's explicit structure (acts, scenes, speeches, lines) and its social layer (characters).
Edges encode both hierarchical inclusion (e.g., $Line \to Speech \to Scene$) and sequential flow.
As shown in Table~\ref{tab:graph_stats}, this topology provides the scaffold onto which latent motifs are attached.

\subsection{The Critical Corpus (Weak Supervision)}

We call the supervision \textit{weak} because the labels are recovered automatically from published
criticism rather than written for this task, which makes them both noisy and positive-only.

To instantiate the motif extraction pipeline of Section~\ref{subsec:instantiation}, we curated a
corpus of \dataNumDocs{} documents spanning from 1765 to 2020.
Crucially, rather than random selection, we stratified sources into three interpretive lenses (Table~\ref{tab:corpus_stratification})
to ensure the narrative graph captures diverse semantic dimensions:

\begin{figure}[t]
    \vspace{-0.8em}
    \centering
    \includegraphics[width=\linewidth]{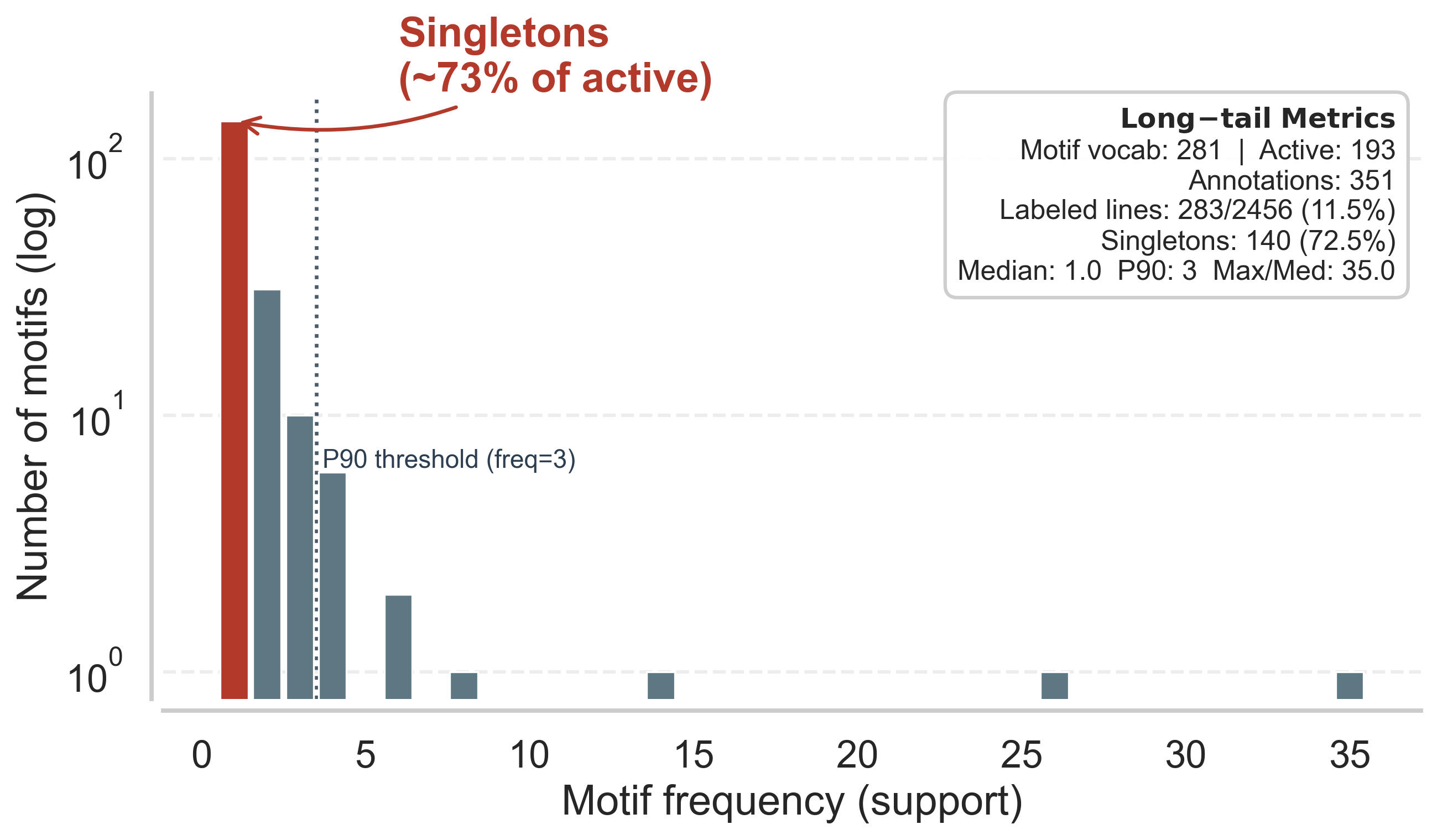}
    \vspace{-0.8em}
    \caption{\textbf{The Severe Long-Tail Distribution of Literary Motifs.}
    \textsc{MotifBench} follows a sharp power law (Imbalance Ratio = \dataImbalanceRatio{}), with over \textbf{96\%} of \textit{active} motifs (\dataTailCount{}/\dataNumActiveMotifs{}) in the sparse ``tail'' (freq $\le$ 5).
    This extreme sparsity renders standard random splits ineffective, motivating our transductive formulation to capture these rare signals.}
    \label{fig:longtail}
    \vspace{-0.8em}
\end{figure}

Crucially, we adopted an \textit{accessibility-first} policy to facilitate reproducibility.
The majority of our corpus is sourced from \textbf{Project Gutenberg} \cite{projectgutenberg} and is in the public domain.
To strictly adhere to intellectual property rights, our public dataset release follows a split-distribution protocol:
we provide the full digitized text for public domain works,
while copyrighted sources (mostly recent phenomenological criticism) are released as structured metadata (\eg DOI, Chapter Title)
to enable researchers to retrieve the source text independently while preserving the integrity of the graph structure.

\subsection{Challenge: The Long-Tail Topology}

A quantitative analysis reveals that literary motifs follow a severe long-tail distribution (Figure \ref{fig:longtail}).
The dataset exhibits an Imbalance Ratio (IR), defined as $\frac{freq_{max}}{freq_{median}}$, of \dataImbalanceRatio{}.
While the mean frequency of active motifs is 1.82, the P90 frequency is only 3.0,
indicating that 90\% of active thematic concepts are supported by three or fewer examples.
Crucially, the tail classes (frequency $\le$ 5) comprise \textbf{\dataTailPct{}} (\dataTailCount{}/\dataNumActiveMotifs{}) of the active vocabulary.

This extreme sparsity validates our transductive formulation:
in a standard random inductive split, the \dataSingletons{} motifs attested by a single line---roughly \dataSingletonPct{} of the active vocabulary---would have zero training examples in the labeled set, resulting in topological isolation.
By treating the problem as transductive link prediction,
\ce enables the propagation of weak signals from the few ``head'' motifs to the sparse ``tail'' via the dense narrative backbone.

%% file: experiment.tex
We evaluate \ce against state-of-the-art LLMs and embedding models on the \textit{Macbeth} motif discovery benchmark.
The task is formulated as \textit{inventory-constrained retrieval}:
given a query unit $u$, the model must rank the fixed motif inventory $\mathcal{M}$ ($|\mathcal{M}|=\dataNumMotifs{}$).
We report \textbf{Hit@$k$}, \textbf{nDCG@$k$} and \textbf{MRR} over $N = \dataNumQueryLines{}$
annotated lines.
Literary criticism is sparse and positive-only. An unannotated line--motif pair may be an overlooked
reading rather than a verified negative, so we cannot decisively treat it as a false positive.
We therefore use rank position as the basis of our primary metrics.
Metric definitions and detailed experimental settings are provided in
Appendix~\ref{app:reproducibility}.

\subsection{Main Results}
\label{subsec:main_results}

Table \ref{tab:main_results} presents the main results.
We compare \ce against three baseline families:
(1) \textbf{Embedding Retrievers} (Sentence-BERT, BGE);
(2) \textbf{Open-Source LLMs} (Llama-3, Qwen, DeepSeek-R1);
and (3) \textbf{Proprietary Foundation Models} (GPT-series, Claude, Gemini-3), representing the upper bound of current general-purpose reasoning.

\paragraph{\ce establishes a new state-of-the-art.}
As shown in Table \ref{tab:main_results}, \ce achieves decisive improvements across all metrics.
Our model achieves a Hit@5 of \textbf{\resCEHitFive{}}, nearly quadrupling the performance of the leading proprietary baseline, Gemini-3-pro-preview (\resGeminiHitFive{}).
Crucially, while proprietary models outperform open weights, they hit a distinct ``semantic ceiling'' in the low-20s range (Hit@5).
This suggests that scaling model parameters or context windows alone cannot resolve the structural dependencies of the \ssjuzet;
the gap remains topological, not computational.

\begin{table*}[!t]
    \vspace{-0.8em}
    \centering
    \small
    \setlength{\tabcolsep}{3pt}
    \renewcommand{\arraystretch}{1.1}
    \begin{tabularx}{\textwidth}{@{}l*{4}{>{\centering\arraybackslash}X}*{4}{>{\centering\arraybackslash}X}>{\centering\arraybackslash}X@{}}
      \toprule
      \textbf{Model Configuration}
      & \multicolumn{4}{c}{\textbf{Hit}}
      & \multicolumn{4}{c}{\textbf{nDCG}}
      & \textbf{MRR}\\
      \cmidrule(lr){2-5}\cmidrule(lr){6-9}
      & @1 & @3 & @5 & @10
      & @1 & @3 & @5 & @10
      & \\
      \midrule
      \textit{w/o pre-training (frozen random)}
      & 0.1237 & 0.2544 & 0.3145 & 0.3746
      & 0.1237 & 0.1710 & 0.1944 & 0.2134
      & 0.2189 \\
      \midrule
      \ce\ (\textit{Hetero-only})
      & 0.2792 & 0.4947 & 0.5901 & 0.7208
      & 0.2792 & 0.3742 & 0.4165 & 0.4657
      & 0.4245 \\
      \ce\ (\textit{Seq-only})
      & 0.4876 & 0.6608 & 0.7562 & 0.8198
      & 0.4876 & 0.5543 & 0.5960 & 0.6223
      & 0.6019 \\
      \textbf{\ce\ (\textit{Full Topology})}
      & \textbf{0.5654} & \textbf{0.7809} & \textbf{0.8481} & \textbf{0.9293}
      & \textbf{0.5654} & \textbf{0.6618} & \textbf{0.6979} & \textbf{0.7321}
      & \textbf{0.6942} \\
      \bottomrule
    \end{tabularx}
    \vspace{-0.8em}
    \caption{\textbf{Structural pre-training and topology ablation} ($N{=}\dataNumQueryLines{}$ query lines; $|\mathcal{M}|{=}\dataNumMotifs{}$ motifs).
    We evaluate with a frozen-encoder probing protocol to measure representation quality.
    Pre-training is necessary (random vs. pre-trained), and integrating richer structure yields monotonic gains
    (\textit{Hetero-only} $<$ \textit{Seq-only} $<$ \textit{Full Topology}).}
    \label{tab:ablation}
\end{table*}

\paragraph{Retrieval augmentation yields diminishing returns.}
We experimented with Self-RAG, providing models with semantically similar lines.
The utility of RAG proves inconsistent: while it aids smaller models and boosts Recall for Gemini-3-pro-preview (Hit@10 rises to 33\%), it frequently degrades the ranking quality (nDCG) for reasoners like Claude and GPT-5.2.
Even for the strongest model, RAG only provides marginal gains ($\sim$1\% boost in Hit@5), failing to close the structural gap.
This confirms that \textbf{lexical similarity is a poor proxy for thematic recurrence}, often introducing competing contexts rather than clarifying the motif.

\subsection{Analysis}
\label{sec:scaling_limits}

Our results reveal that model scale alone offers an inefficient path to structural understanding, manifesting in two distinct failure modes.

\paragraph{Inverse Scaling in Instruction Following.}
Counter-intuitively, the largest open model, \textit{GPT-OSS-120B}, underperforms its smaller 20B counterpart (zero-shot Hit@5: 0.099 vs. 0.103).
Error analysis links this to \textbf{schema non-compliance}: the 120B model prioritizes generative fluency over strict output constraints.
This confirms the ``inverse scaling'' phenomenon \cite{mckenzie2023inverse}, where strong generative priors actively hinder rigid classification tasks.

\paragraph{The ``Semantic Ceiling'' and Efficiency Paradox.}
Comparing open weights to proprietary giants (e.g., Gemini-3), we observe that while scaling provides gains (Hit@5 rising from $\sim$\resOpenCeilingHitFive{} to $\sim$\resPropCeilingHitFive{}),
it hits sharp diminishing returns.
Despite massive compute and likely \textbf{positive data contamination} (exposure to critical texts),
general-purpose models hit a distinct ``Ceiling'' in the low-20s range.
In contrast, \ce demonstrates a nearly $4\times$ performance improvement (\resCEHitFive{}) with only a fraction of the parameter count.
This proves that for narrative topology, \textbf{explicit structural priors} are a far more efficient inductive bias than brute-force scaling,
which can only asymptotically approximate the \ssjuzet.

\subsection{Ablation Study}
\label{subsec:ablation}

To rigorously disentangle the contribution of structural priors from the capacity of the GNN backbone,
we adopt a frozen-encoder probing protocol.
Instead of full fine-tuning, we freeze the pre-trained encoder and optimize only a lightweight matching head.
This strict setup measures how easily motif relevance is \emph{recoverable} from the representation space itself.

\paragraph{Structure-First Pre-training is Essential.}
Table \ref{tab:ablation} reveals a stark separation in representation quality.
With a randomly initialized encoder, performance is poor (Hit@5 $\approx$ \resAblRandomHitFive{}), implying that motif relevance is not intrinsic to the raw graph.
In contrast, our pre-trained encoder achieves Hit@5 $\approx$ \resAblFullHitFive{}.
This absolute gain of \textbf{\resAblPretrainGain{}} confirms that the self-supervised objectives (Section \ref{subsec:transfer})
effectively transform raw narrative connectivity into a semantically aligned space,
making motif discovery a linear problem rather than a complex non-linear one.

\paragraph{Synergy of Heterogeneous Topology.}
We further ablate the topological components (Table \ref{tab:ablation}). A clear hierarchy emerges:
\begin{enumerate}
    \item \textit{Hetero-only} ($\mathcal{E}_{\textsc{hier}} \cup \mathcal{E}_{\textsc{agent}}$; Hit@5 \resAblHeteroHitFive{}) provides useful global anchors (e.g., character/scene associations) but lacks local context.
    \item \textit{Seq-only} ($\mathcal{E}_{\textsc{seq}}$; Hit@5 \resAblSeqHitFive{}) captures the local discourse flow, which is information-rich but locally ambiguous.
    \item \textbf{\textit{Full Topology}} ($\mathcal{E}_{\textsc{seq}} \cup \mathcal{E}_{\textsc{hier}} \cup \mathcal{E}_{\textsc{agent}}$; Hit@5 \resAblFullHitFive{}) achieves the best performance.
\end{enumerate}
This result confirms a complementary effect:
heterogeneous anchors (e.g., \textit{SpeakerOf}) help disambiguate locally similar utterances that play different roles in the dramatic scaffold.

\begin{figure}[!t]
    \vspace{-0.8em}
    \centering
    \includegraphics[width=0.4\textwidth]{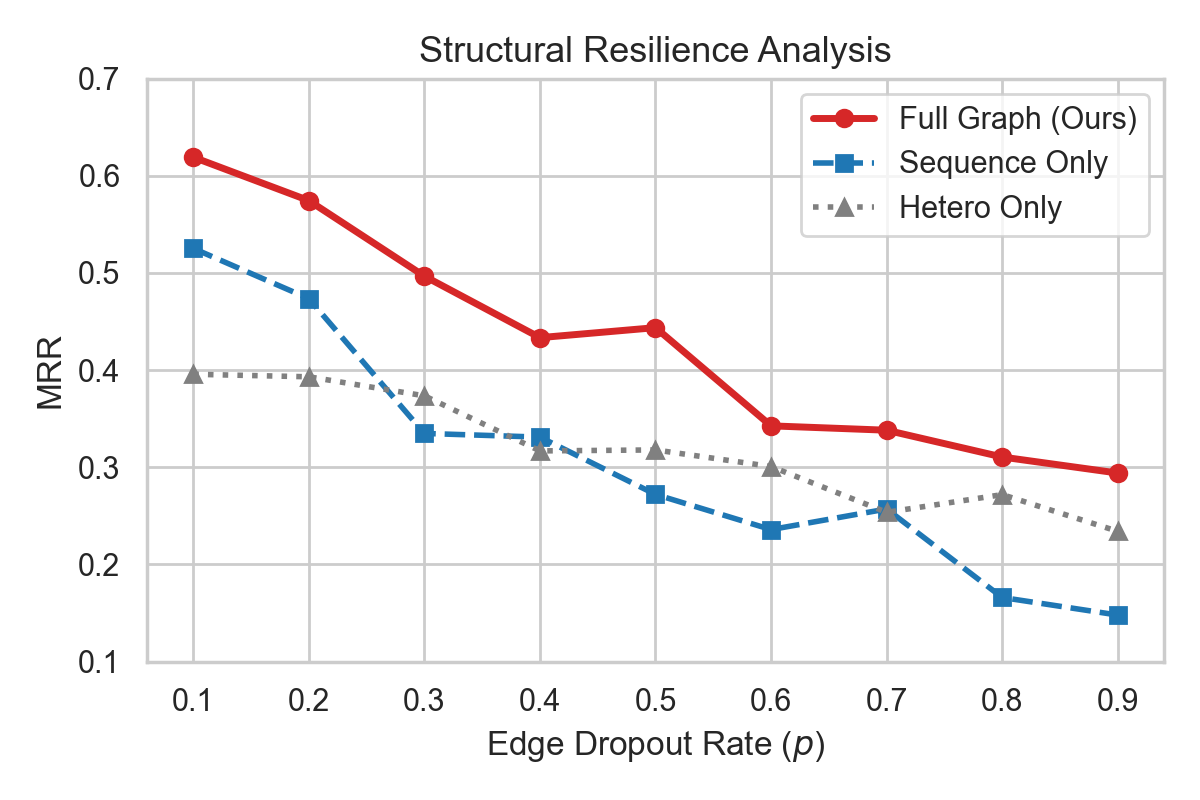}
    \vspace{-0.8em}
    \caption{\textbf{Structural resilience under edge dropout.}
    Mean Reciprocal Rank (MRR) as a function of inference-time edge dropout rate $p$.
    The Full topology degrades more gracefully than Sequence-only across a wide range of dropout rates,
    indicating higher robustness to partial graph observations.}
    \label{fig:resilience}
    \vspace{-0.8em}
\end{figure}

\paragraph{Structural Resilience and Redundancy.}
Beyond peak accuracy, we stress-test robustness by applying random edge dropout at inference time (Figure \ref{fig:resilience}).
As the dropout rate $p$ increases, the \textit{Sequence-only} model degrades sharply, reflecting its fragility to broken local contexts.
In contrast, the \textit{Full} topology maintains consistently higher MRR and exhibits a smoother decline.
This demonstrates \textbf{structural redundancy}:
when the local sequential path is severed, the heterogeneous edges ($\mathcal{E}_{\textsc{hier}}, \mathcal{E}_{\textsc{agent}}$) provide alternative message-passing routes,
preserving thematic signals through the "dilatory space" of the narrative.

\subsection{Qualitative Analysis: Structural Disambiguation}
\label{subsec:case_study}

\begin{figure}[!t]
    \centering
    \includegraphics[width=0.5\textwidth]{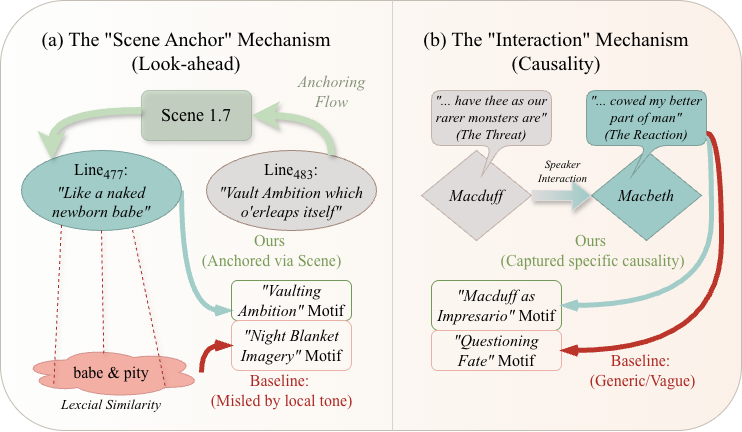}
    \vspace{-0.8em}

    \caption{\textbf{Structural reasoning beyond local cues.}
    (a) \ce anchors a metaphor to its scene context, avoiding a lexical-similarity trap.
    (b) \ce propagates cross-speaker causality via interaction structure, resolving a specific motif where a sequential baseline predicts a generic one.}

    \label{fig:case_studies}
    \vspace{-0.8em}
\end{figure}

Figure~\ref{fig:case_studies} demonstrates how topological priors resolve ambiguities that mislead sequence-only models.

\noindent\textbf{Case (a): Scene Anchoring.}
The line \textit{``pity, like a naked newborn babe''} (Act~1, Scene~7) contains salient lexical imagery that distracts sequence baselines toward \textit{Night Imagery}.
\ce instead routes evidence through the \texttt{Scene} node, aggregating thematically consistent cues within the same dramatic unit.
In particular, Act~1, Scene~7 also contains the explicit ambition framing (e.g., ``vaulting ambition'', Line~\#483), making \textbf{\textit{Ambition}} topologically supported even when the target span itself is metaphorical.

\noindent\textbf{Case (b): Interactional Causality.}
For Macbeth's reaction \textit{``it hath cowed my better part of man''}, baselines default to an underspecified \textit{Questioning Fate} label when encoding the line in isolation.
Our graph makes the immediate antecedent accessible via the \texttt{spoke\_by} edge from Macduff's threat to exhibit Macbeth as a ``monster'',
allowing \ce to recover the specific \textbf{\textit{Impresario}} motif grounded in the cross-speaker exchange.

\paragraph{Why topology helps.}
Both cases illustrate \emph{structural under-conditioning} in sequence encoders: the decisive evidence is not in the target span but in its structural neighborhood (within the same scene or an adjacent turn).
In our graph, such cues become reachable via short typed paths (Line$\!\rightarrow\! $Scene$ \rightarrow $Line; Line$\rightarrow\texttt{spoke\_by}\rightarrow$Character), imposing an explicit \emph{evidence-routing prior}.
As a result, \ce prefers motifs supported by coherent structure over those driven by local lexical salience or underspecified global labels. Detailed analysis is provided in Appendix~\ref{app:case_studies}.

%% file: conclusion.tex
We presented \ce, a graph-based framework that bridges the gap between the linear \ffabula and the topological \ssjuzet in narrative understanding.
While standard sequential models struggle with the non-chronological distortions of literary discourse,
our framework operationalizes these structural dependencies into a heterogeneous graph.
By employing a transductive learning objective,
our approach overcomes the severe long-tail sparsity of thematic motifs, significantly outperforming LLM baselines.
This indicates that the introduced structural inductive bias provides necessary scaffolding for reasoning about dispersed narrative clues.
Consequently, our work suggests that for deep literary analysis, models must move beyond sequence modeling to internalize the complex, non-linear architecture of storytelling.

%% file: limit.tex
While our results demonstrate the efficacy of structure-aware pre-training,
we acknowledge inherent constraints regarding granularity, data density, and the trade-off between precision and scalability.

\paragraph{Genre Adaptation and Granularity.}
Our experiments focus on drama, where topological markers (Acts/Scenes/Turns) are explicit.
Generalizing to prose fiction requires fundamentally different, finer-grained annotation strategies.
In novels, the \ssjuzet organization is often implicit (e.g., free indirect discourse, stream of consciousness) and lacks clear segmentation.
Adapting \ce to such forms would necessitate computationally expensive discourse parsing to recover the latent scaffold,
imposing a trade-off between structural precision and the breadth of compatible literary forms.

\paragraph{The ``Well-Studied'' Assumption.}
Our pipeline operates under a \textbf{canonical work assumption}:
it relies on the availability of a rich scholarly infrastructure---
specifically, high-fidelity structural backbones (TEI) and dense expert commentary.
This restricts our method to ``well-studied'' texts (e.g., Shakespeare, Joyce) where such dual-layer supervision is available.
Applying this framework to ``in-the-wild'' web fiction or under-resourced languages would require robust upstream parsing,
where noise could propagate to the GNN and degrade representation quality.

\paragraph{Closed-World Constraint vs. Open Discovery.}
We currently frame motif identification as retrieval over a fixed, expert-curated inventory ($|\mathcal{M}| = \dataNumMotifs{}$).
This closed-world assumption allows for high-precision alignment with scholarly consensus but precludes \textit{ab initio} discovery of novel or evolving themes.
Unlike self-supervised LLMs that consume raw text to maximize generality, \ce serves as a \textbf{specialized, high-precision instrument} for Deep Digital Humanities.
While this design ensures rigorous modeling of complex narrative architectures,
it inherently limits the system's ability to scale as a general-purpose foundation model for arbitrary text analysis.

\paragraph{Linguistic Domain Shift.}
Our node initialization relies on off-the-shelf sentence encoders (e.g., BGE) trained primarily on modern English.
This introduces a \textbf{domain gap} when processing the Early Modern English of Shakespeare (e.g., archaic syntax and semantic drift).
While our structural pre-training objectives partially mitigate this by contextualizing embeddings within the narrative topology,
the underlying semantic representations remain suboptimal compared to a model linguistically adapted to the specific historical period.

\paragraph{Downstream Integration with Generative Models.}
Our current framework focuses on the \textit{discriminative} task of motif discovery.
While we establish that structural priors are essential for accurate retrieval, we leave the question of \textbf{downstream utilization} under-explored.
Specifically, how these structure-aware topological embeddings can be injected back into LLMs---
for instance, to guide controllable text generation, automate literary essay writing, or enhance Graph-RAG pipelines---
remains an open research direction.
We envision \ce not as a replacement for foundation models,
but as a modular structural adapter that bridges the gap between raw text processing and deep thematic reasoning.

%% file: ethics.tex
\paragraph{Data Usage and Intellectual Property.}
The primary textual data used in this study, William Shakespeare's \textit{Macbeth}, is in the public domain.
The critical essays and secondary literature used for motif inventory construction were accessed
through academic repositories consistent with fair use principles for research purposes.
We have ensured that our released dataset contains no personally identifiable information (PII) or copyright-infringing material.

\paragraph{Content Safety and Literary Context.}
While the source text (\textit{Macbeth}) inherently contains themes of violence, regicide, and psychological distress,
these are analyzed strictly within a literary and historical context.
In our use of Large Language Models (LLMs) for weak supervision and comparative evaluation,
we implemented prompt constraints to ensure the models remained focused on literary analysis and did not generate gratuitous, harmful, or non-consensual content.
We manually reviewed a subset of model outputs to verify adherence to these safety guidelines.

\paragraph{Model Usage and Compute.}
We utilized both open-weights models (e.g., Llama-3, Qwen) and proprietary APIs (e.g., GPT, Claude).
We adhered to the respective Terms of Use and Acceptable Use Policies of all service providers.
Furthermore, our proposed method (\ce) relies on lightweight Graph Neural Networks,
which incur significantly lower computational costs and carbon footprint compared to fine-tuning Large Language Models,
aligning with Green AI principles.

\paragraph{Human Subjects.}
This research did not involve human subjects or crowdsourced annotators (e.g., Amazon Mechanical Turk).
All "ground truth" labels were derived from established literary criticism or expert verification by the authors.
Therefore, no Institutional Review Board (IRB) approval was required.

%% file: availability.tex
Code is released at \url{https://github.com/Acture/critics-eye} (AGPL-3.0); the \textsc{MotifBench} dataset (extracted line-motif tuples, motif descriptors, Folger Through-Line-Number anchors, and Folger line text) is released separately at \url{https://huggingface.co/datasets/Acture/MotifBench} (CC~BY-NC~4.0), so that code and data can be versioned and cited independently. The dataset's license follows from its content: it incorporates the Folger Shakespeare Library's digital edition of \textit{Macbeth}, licensed CC~BY-NC~3.0, so the release as a whole cannot be more permissive than that source. Consistent with the split-distribution protocol described in Section~\ref{sec:dataset}, public-domain criticism sources are pointed to in full and copyrighted sources are released as bibliographic metadata only.

%% file: appendix/corpus.tex
The detailed list is shown in Table~\ref{tab:corpus_details} on page~\pageref{tab:corpus_details}.

\begin{table*}[t]
	\centering
	\caption{The Critical Corpus for \textit{Macbeth} Motif Extraction (Weak Supervision Sources).}
	\label{tab:corpus_details}

	\small
	\setlength{\tabcolsep}{4pt}
	\renewcommand{\arraystretch}{1.2}

	\begin{tabularx}{\textwidth}{@{} p{0.25\textwidth} >{\raggedright\arraybackslash}X @{}}
		\toprule
		\textbf{Author / Source}         & \textbf{Title}                                                                                                                     \\
		\midrule

		\citet{johnson1765preface}       & \textit{Preface to Shakespeare}                                                                                                    \\
		\citet{fletcher1844macbeth}      & \textit{Macbeth:---Shakespearian Criticism and Acting}                                                                               \\
		\citet{weiss1876wit}             & \textit{Wit, Humor, and Shakspeare: Twelve Essays}                                                                                 \\
		\citet{dyer1883folklore}         & \textit{Folk-lore of Shakespeare}                                                                                                  \\
		\citet{hazlitt1903macbeth}       & \textit{Characters of Shakespeare's Plays}                                                                                  \\
		\citet{bradley1904shakespearean} & \textit{Shakespearean Tragedy: Lectures on Hamlet, Othello, King Lear, Macbeth}                                                    \\
		\citet{tolstoy1906shakespeare}   & \textit{Tolstoy on Shakespeare: A Critical Essay on Shakespeare}                                                                   \\
		\citet{adams1917playhouses}      & \textit{Shakespearean Playhouses: A History of English Theatres from the Beginnings to the Restoration}                            \\
		\citet{croce1920ariosto}         & \textit{Ariosto, Shakespeare and Corneille}                                                                                        \\
		\citet{herford1921treatment}     & \textit{Shakespeare's Treatment of Love \& Marriage, and Other Essays}                                                             \\
		\citet{smith1961shakespeare}     & \textit{Shakespeare Criticism: A Selection, 1623-1840}                                                                             \\
		\citet{gerlach1973shakespeare}   & \textit{Shakespeare, Kurosawa, and "Macbeth": A Response to J. Blumenthal}                                                         \\
		\citet{cheung1984dread}          & \textit{Shakespeare and Kierkegaard: "Dread" in Macbeth}                                                                           \\
		\citet{bloom1987william}         & \textit{William Shakespeare's Macbeth}                                                                                             \\
		\citet{kishi1994shakespeare}     & \textit{Shakespeare and Cultural Traditions: The Selected Proceedings of the International Shakespeare Association World Congress} \\
		\citet{keller1995moral}          & \textit{The Moral Thinking of Macbeth}                                                                                             \\
		\citet{harris2007smell}          & \textit{The Smell of Macbeth}                                                                                                      \\
		\citet{bloom2008macbeth_ages}    & \textit{Macbeth (Bloom's Shakespeare Through the Ages)}                                                                            \\
		\citet{moschovakis2008macbeth}   & \textit{Macbeth: New Critical Essays}                                                                                              \\
		\citet{spolsky2011embodied}      & \textit{An Embodied View of Misunderstanding in Macbeth}                                                                           \\
		\citet{cox2013religion}          & \textit{Religion and Suffering in Macbeth}                                                                                         \\
		\citet{sigit2013tragedy}         & \textit{Tragedy and Moral Values in William Shakespeare's Macbeth: A Structural Analysis}                                          \\
		\citet{omura2018sympathy}        & \textit{The Problem of Sympathy in Macbeth}                                                                                        \\
		\citet{muttaleb2019political}    & \textit{Macbeth's Political Imagination: The Struggle for Kingship in Macbeth}                                                     \\
		\citet{sachon2020shakespeare}    & \textit{Shakespeare, Objects and Phenomenology: Daggers of the Mind}                                                               \\
		\citet{ignoumacbethunit4}        & \textit{Unit 4: Macbeth: Critical Responses}                                                                                       \\

		\bottomrule
	\end{tabularx}
\end{table*}

%% file: appendix/prompts.tex
This appendix gives the two prompts used in the paper. The first builds the supervision signal from
criticism (Section~\ref{subsec:instantiation}). The second is the zero-shot ranking prompt for the
LLM baselines (Section~\ref{subsec:main_results}).

\subsection{Relation Extraction Prompt}

Extraction uses GPT-OSS-120B at \texttt{temperature=0.0} behind a Pydantic schema. The interaction
consists of a system instruction defining the critical persona, a user query containing the target
passage, and an iterative repair mechanism to ensure valid JSON output.

\begin{figure}[h!]
    \centering
    \small
    \begin{tcolorbox}[colback=gray!10!white, colframe=black!75!black, title=\textbf{System Prompt (Relation Extraction)}]
        You are a literary analysis assistant specializing \textit{only} in William Shakespeare's
        play \textit{Macbeth}.

        \vspace{0.5em}
        \textbf{Prerequisites.} Return \texttt{[]} unless all three hold: \\
        (1) the sentence states a relationship between two distinct elements \textit{within} the
        play, whether literary (foreshadowing, parallelism, symbolism) or structural; \\
        (2) it carries at least one citation to the play in \texttt{number.number.number} form
        (for instance \texttt{1.7.16}). Page numbers, years, and citations to other works do not qualify; \\
        (3) the relationship is stated by the critic, not a description of the critic's own argument
        or of general theory.

        \vspace{0.5em}
        \textbf{Fields:} \texttt{source\_element}, \texttt{target\_element},
        \texttt{relationship\_type} (a JSON list drawn from a fixed label set),
        \texttt{source\_citation} and \texttt{target\_citation} as \texttt{[Act, Scene, Line]} with
        \texttt{null} for any missing component, \texttt{justification\_quote}, \texttt{confidence}.
    \end{tcolorbox}

    \vspace{0.2cm}

    \begin{tcolorbox}[colback=gray!10!white, colframe=black!75!black, title=\textbf{User Prompt}]
        \texttt{Extract relations from the following text: [criticism passage]}
    \end{tcolorbox}

    \vspace{0.2cm}

    \begin{tcolorbox}[colback=red!5!white, colframe=red!75!black, title=\textbf{Error Correction Prompt (Triggered on Parse Fail)}]
        Your last output was invalid. Please fix it. \\
        \textbf{Error:} \texttt{[validator message]} \\
        \textbf{Bad output:} \texttt{[model output]}
    \end{tcolorbox}

    \caption{Prompt templates for relation extraction from criticism
    (Section~\ref{subsec:instantiation}), condensed. The full prompt also carries four worked
    examples, two of which show passages that must be rejected.}
    \label{fig:extraction_prompts}
\end{figure}

\subsection{Output Schema (Extraction)}

Generation is constrained by the following Pydantic schema. Output that fails validation is returned
to the model once with the validator message, and is discarded if it fails again.

\begin{quote}
\small
\begin{verbatim}
{
  "source_element": "string",
  "target_element": "string",
  "relationship_type": ["label", ...],
  "source_citation": [Act, Scene, Line],
  "target_citation": [Act, Scene, Line],
  "justification_quote": "string",
  "confidence": float
}
\end{verbatim}
\end{quote}

\subsection{Filtering and Retention}

Three filters stand between a criticism document and a supervision edge. Before generation, each
document is split into 2{,}000-character chunks, and a chunk reaches the model only if it contains a
Shakespeare-style citation and either names the play or carries one of the allowed relation labels.
At generation, output must satisfy the schema above. At grounding, a tuple becomes an edge only
where the critic's own \textit{(Act, Scene, Line)} citation resolves to a node of the parsed play.

Extraction over the \dataNumDocs{}-document corpus produced 365 tuples. Each tuple supplies its
source and target element as candidate descriptors, so 730 citations were available to resolve.

\begin{center}
\small
\begin{tabular}{@{}lr@{}}
\toprule
\textbf{Citation outcome} & \textbf{Count} \\
\midrule
Resolves to a line              & 367 \\
Resolves to a speech section    & 97 \\
Resolves to a scene             & 2 \\
No citation supplied            & 249 \\
Supplied but unresolvable       & 15 \\
\midrule
\textbf{Total}                  & \textbf{730} \\
\bottomrule
\end{tabular}
\end{center}

Of the 481 citations supplied, 466 resolve, and 355 of the 365 tuples retain at least one resolvable
endpoint. The 367 line-resolving descriptors yield \dataNumAnnotations{} distinct line--motif edges
once repeated assertions of the same pair are merged. Every retained edge therefore carries a
citation the critic supplied, and a claim made without one cannot enter the benchmark.

\subsection{Motif Ranking Prompt (LLM Baselines)}

For the zero-shot LLM baselines we supply the target line and the full inventory of
\dataNumMotifs{} motifs with their labels, and the model returns a ranked list of motif IDs. This is
a separate procedure from the extraction pipeline above.

\begin{figure}[h!]
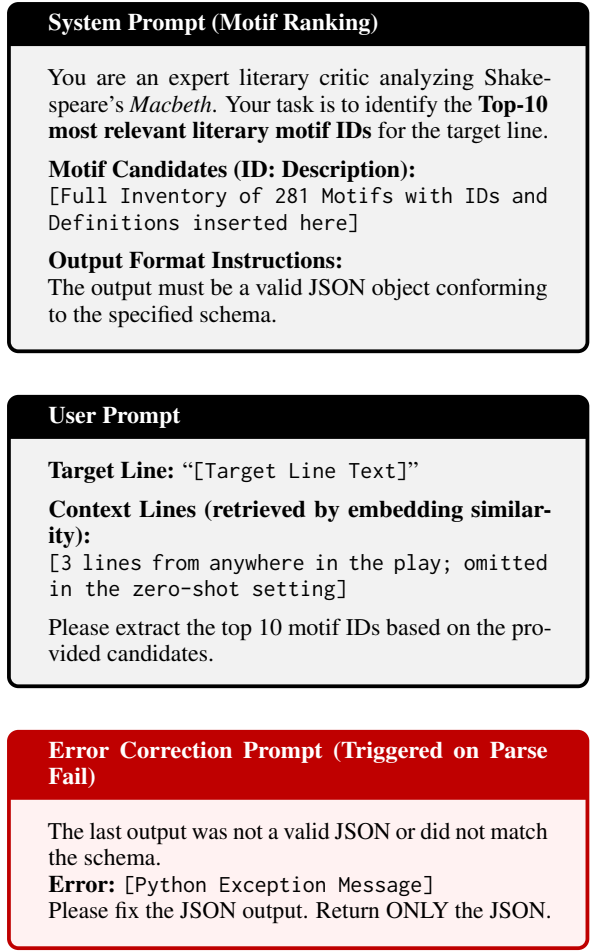

    \centering
    \small
    \begin{tcolorbox}[colback=gray!10!white, colframe=black!75!black, title=\textbf{System Prompt (Motif Ranking)}]
        You are an expert literary critic analyzing Shakespeare's \textit{Macbeth}.
        Your task is to identify the \textbf{Top-10 most relevant literary motif IDs} for the target line.

        \vspace{0.5em}
        \textbf{Motif Candidates (ID: Description):} \\
        \texttt{[Full Inventory of \dataNumMotifs{} Motifs with IDs and Definitions inserted here]}

        \vspace{0.5em}
        \textbf{Output Format Instructions:} \\
        The output must be a valid JSON object conforming to the specified schema.
    \end{tcolorbox}

    \vspace{0.2cm}

    \begin{tcolorbox}[colback=gray!10!white, colframe=black!75!black, title=\textbf{User Prompt}]
        \textbf{Target Line:} ``\texttt{[Target Line Text]}''

        \vspace{0.5em}
        \textbf{Context Lines (retrieved by embedding similarity):} \\
        \texttt{[3 lines from anywhere in the play; omitted in the zero-shot setting]}

        \vspace{0.5em}
        Please extract the top 10 motif IDs based on the provided candidates.
    \end{tcolorbox}

    \vspace{0.2cm}

    \begin{tcolorbox}[colback=red!5!white, colframe=red!75!black, title=\textbf{Error Correction Prompt (Triggered on Parse Fail)}]
        The last output was not a valid JSON or did not match the schema. \\
        \textbf{Error:} \texttt{[Python Exception Message]} \\
        Please fix the JSON output. Return ONLY the JSON.
    \end{tcolorbox}

    \caption{Prompt templates for the zero-shot and retrieval-augmented LLM baselines
    (Section~\ref{subsec:main_results}). This is a separate procedure from the extraction pipeline
    of Figure~\ref{fig:extraction_prompts}.}
    \label{fig:baseline_prompts}
\end{figure}

\subsection{Output Schema (Ranking)}
We enforced structured generation using a Pydantic definition to guarantee that the output is parsable. The schema requires a ranked list of integers.

\begin{quote}
\begin{verbatim}
{
  "motif_ids": [
    integer,  // Rank 1 Motif ID
    integer,  // Rank 2 Motif ID
    ...
    integer   // Rank 10 Motif ID
  ]
}
\end{verbatim}
\end{quote}

%% file: appendix/reproducibility.tex
We detail the hyperparameters and hardware configurations to ensure reproducibility.
\subsection{Hyperparameters}
\begin{itemize} [noitemsep, topsep=0pt]
	\item \textbf{Encoder Architecture:} 2-layer Heterogeneous Graph Transformer (HGT).
	\item \textbf{Hidden Dimension:} 768.
	\item \textbf{Optimization:} AdamW optimizer with learning rate 2e-5, weight decay 0.01.
	\item \textbf{Batch Size / Epochs:} 32 / 50, on a single Apple M2 (macOS; PyTorch MPS).
	\item \textbf{Compute Budget:} training completed within $<1$ day wall-clock time on the above machine.
\end{itemize}
\subsection{Baseline Configurations}
For closed-source models (e.g., GPT-5.2-Chat, Claude-Opus-4.5), we utilized OpenRouter's respective APIs with temperature set to 0.0 to maximize determinism.

\subsection{Evaluation Metrics}
One query is one annotated line $u$, of which there are $N = \dataNumQueryLines{}$. Its gold set
$G(u) \subseteq \mathcal{M}$ is every motif that criticism attaches to $u$. A system ranks the whole
inventory $\mathcal{M}$ ($|\mathcal{M}| = \dataNumMotifs{}$); write $r_u(m)$ for the 1-based rank of
motif $m$ and $\pi_u(i)$ for the motif at rank $i$. All three measures are averaged over the $N$
queries.

\begin{itemize} [noitemsep, topsep=0pt]
	\item \textbf{Hit@$k$} is $1$ when $\min_{m \in G(u)} r_u(m) \le k$ and $0$ otherwise. It asks
	whether a shortlist of length $k$ contains anything the critics recorded, and is indifferent to
	position within that shortlist.
	\item \textbf{MRR} is $1 / \min_{m \in G(u)} r_u(m)$. It measures how far down the ranking a
	reader must go to reach the first correct motif, and so is sensitive to position where Hit@$k$
	is not. It sees only the best-ranked gold motif.
	\item \textbf{nDCG@$k$} is $\mathrm{DCG@}k / \mathrm{IDCG@}k$ under binary gain, with
	$\mathrm{DCG@}k = \sum_{i \le k} \mathbf{1}\{\pi_u(i) \in G(u)\} / \log_2(i+1)$ and
	$\mathrm{IDCG@}k = \sum_{i \le \min(k, |G(u)|)} 1 / \log_2(i+1)$. It is the only one of the
	three that rewards ranking \emph{all} of a line's motifs highly, and the normalization keeps a
	line carrying several motifs comparable with one carrying a single motif.
\end{itemize}

The three are complementary because a line can participate in more than one motif. Hit@$k$ asks
whether the shortlist is usable at all, MRR asks how quickly it becomes usable, and nDCG@$k$ asks
whether the rest of that line's motifs were recovered too.

%% file: appendix/fullresult.tex
Table \ref{tab:full_results} presents the comprehensive performance metrics for all evaluated models,
including smaller open-source baselines (e.g., Qwen-4B) and random guess baselines omitted from the main text for brevity.
\begin{table*}[t]
    \centering
    \small
    \setlength{\tabcolsep}{3pt}
    \renewcommand{\arraystretch}{1.1}

    \begin{tabularx}{\textwidth}{@{}l*{4}{>{\centering\arraybackslash}X}*{4}{>{\centering\arraybackslash}X}>{\centering\arraybackslash}X@{}}
    \toprule
    \textbf{Method}
    & \multicolumn{4}{c}{\textbf{Hit}}
    & \multicolumn{4}{c}{\textbf{nDCG}}
    & \textbf{MRR}\\
    \cmidrule(lr){2-5}\cmidrule(lr){6-9}
    & @1 & @3 & @5 & @10
    & @1 & @3 & @5 & @10
    & \\

    \midrule
    \multicolumn{10}{l}{\textit{\textbf{Embedding Baselines}}} \\
    \quad Random Guess
    & 0.0035 & 0.0106 & 0.0177 & 0.0353
    & 0.0035 & 0.0075 & 0.0104 & 0.0161
    & 0.0220 \\
    \quad Sentence-BERT
    & 0.0424 & 0.0742 & 0.1060 & 0.1060
    & 0.0406 & 0.0531 & 0.0637 & 0.0736
    & 0.0812 \\
    \quad BGE-Large-v1.5
    & 0.0389 & 0.0707 & 0.0813 & 0.1307
    & 0.0371 & 0.0488 & 0.0521 & 0.0665
    & 0.0748 \\

    \midrule
    \multicolumn{10}{l}{\textit{\textbf{Open-Source LLMs}}} \\
    \textit{Qwen-3-4B} & & & & & & & & & \\
    \quad Zero-shot
    & 0.0247 & 0.0353 & 0.0459 & 0.0565
    & 0.0247 & 0.0305 & 0.0349 & 0.0384
    & 0.0328 \\
    \quad + RAG
    & 0.0212 & 0.0318 & 0.0636 & 0.0742
    & 0.0212 & 0.0279 & 0.0413 & 0.0446
    & 0.0354 \\

    \textit{Llama-3-8B} & & & & & & & & & \\
    \quad Zero-shot
    & 0.0318 & 0.0707 & 0.0777 & 0.0989
    & 0.0318 & 0.0545 & 0.0574 & 0.0641
    & 0.0531 \\
    \quad + RAG
    & 0.0283 & 0.0459 & 0.0459 & 0.0530
    & 0.0283 & 0.0380 & 0.0380 & 0.0405
    & 0.0364
    \\

    \textit{GPT-OSS-20B} & & & & & & & & & \\
    \quad Zero-shot
    & 0.0424 & 0.0707 & 0.1025 & 0.1060
    & 0.0424 & 0.0598 & 0.0732 & 0.0742
    & 0.0639 \\
    \quad + RAG
    & 0.0424 & 0.0883 & 0.1166 & 0.1201
    & 0.0424 & 0.0700 & 0.0811 & 0.0822
    & 0.0699 \\

    \textit{DeepSeek-R1-70B} & & & & & & & & & \\
    \quad Zero-shot
    & 0.0813 & 0.1237 & 0.1343 & 0.1590
    & 0.0813 & 0.1057 & 0.1101 & 0.1180
    & 0.1052 \\
    \quad + RAG
    & 0.0742 & 0.1166 & 0.1343 & 0.1802
    & 0.0742 & 0.0991 & 0.1063 & 0.1213
    & 0.1032 \\

    \textit{GPT-OSS-120B} & & & & & & & & & \\
    \quad Zero-shot
    & 0.0247 & 0.0495 & 0.0989 & 0.0989
    & 0.0247 & 0.0394 & 0.0596 & 0.0596
    & 0.0471 \\
    \quad + RAG
    & 0.0318 & 0.0636 & 0.0848 & 0.0848
    & 0.0318 & 0.0491 & 0.0582 & 0.0582
    & 0.0495 \\

    \midrule
    \multicolumn{10}{l}{\textit{\textbf{Closed-Source LLMs (Proprietary)}}} \\

    \textit{GPT-5.2-Chat} & & & & & & & & & \\
    \quad Zero-shot
    & 0.0636 & 0.1272 & 0.1696 & 0.1837
    & 0.0636 & 0.0986 & 0.1155 & 0.1201
    & 0.0998 \\
    \quad + RAG
    & 0.0495 & 0.1166 & 0.1413 & 0.2120
    & 0.0495 & 0.0872 & 0.0974 & 0.1209
    & 0.0928 \\

    \textit{Claude-Opus-4.5} & & & & & & & & & \\
    \quad Zero-shot
    & 0.0777 & 0.1484 & 0.1979 & 0.2756
    & 0.0777 & 0.1177 & 0.1381 & 0.1633
    & 0.1290 \\
    \quad + RAG
    & 0.0671 & 0.1343 & 0.1661 & 0.2615
    & 0.0671 & 0.1058 & 0.1193 & 0.1499
    & 0.1162 \\

    \textit{Grok-4.1-Fast} & & & & & & & & & \\
    \quad Zero-shot
    & 0.0636 & 0.1378 & 0.1731 & 0.2509
    & 0.0636 & 0.1049 & 0.1190 & 0.1439
    & 0.1114 \\
    \quad + RAG
    & 0.0777 & 0.1449 & 0.1802 & 0.2650
    & 0.0777 & 0.1164 & 0.1304 & 0.1572
    & 0.1248
    \\

    \textit{Gemini-3-pro-preview} & & & & & & & & & \\
    \quad Zero-shot
    & 0.0707 & \underline{0.1625} & 0.2155 & 0.2862
    & 0.0707 & 0.1231 & 0.1453 & 0.1683
    & 0.1317 \\
    \quad + RAG
    & \underline{0.0883} & 0.1555 & \underline{0.2261} & \underline{0.3322}
    & \underline{0.0883} & \underline{0.1288} & \underline{0.1562} & \underline{0.1909}
    & \underline{0.1484}
    \\

    \midrule
    \textbf{\ce (Ours)}
    & \textbf{0.5477} & \textbf{0.7739} & \textbf{0.8481} & \textbf{0.9223}
    & \textbf{0.5477} & \textbf{0.6462 } & \textbf{0.6820 } & \textbf{0.7156}
    & \textbf{0.6762} \\

    \bottomrule
    \end{tabularx}

    \begin{minipage}{\textwidth}
        \vspace{1mm}
        \footnotesize
        \textit{Note:} nDCG@1 is equivalent to Hit@1 under binary relevance.
        \textbf{Baselines:} We underline the second-best performance (best among baselines).
        Newer closed-source models (e.g., Claude-Opus-4.5) outperform open weights but still lag significantly behind \ce.
    \end{minipage}

    \caption{Performance on motif identification ($N=\dataNumQueryLines{}$). Comparison against Open and Closed SOTA LLMs.}
    \label{tab:full_results}
\end{table*}

%% file: appendix/casestudy.tex
In this section, we expand the two case studies in Section~\ref{subsec:case_study} by making explicit
(i) what the baselines attend to, (ii) what structural evidence \ce has access to, and
(iii) how this changes the predicted motif.

\subsection{Case Study 1: Metaphor disambiguation via scene-level structure}

\noindent\textbf{Target Line:} \textit{"And pity, like a naked newborn babe"} (Line \#477, Act 1, Scene 7).

\paragraph{Context.}
The line occurs in Macbeth's soliloquy weighing the assassination of Duncan.
Although the local surface contains innocence-related imagery (``babe'') and affect (``pity''),
the surrounding speech in Act~1, Scene~7 foregrounds Macbeth's motive of \textit{ambition}.

\paragraph{Baseline Failures.}
\begin{itemize} [noitemsep, topsep=0pt]
    \item The \textit{Sequence-Only} model misclassifies this as \textit{"Night Blanket Imagery"},
    misled by the somber, metaphorical tone of the immediate text.
    \item The \textit{Heterogeneous-Only} model predicts the generic \textit{"Double Trust in Messengers"},
    relying on coarse character correlations with weak textual grounding.
\end{itemize}

\paragraph{\ce prediction and supporting structure.}
\ce ranks \textbf{\textit{Vaulting Ambition}} as the top motif.
Notably, the explicit phrase ``vaulting ambition'' appears a few lines later (L\#483),
but both lines are located in the same structural unit (Act~1, Scene~7).
In our graph, the target line is connected to the corresponding \texttt{Scene} node,
which in turn links to multiple ambition-related cues within the same scene.
This scene-level connectivity provides a short topological path from the target metaphor to ambition-centered evidence,
reducing the tendency to overfit to local lexical similarity.

\subsection{Case Study 2: Causality across speakers via interaction structure}

\noindent\textbf{Target Line:} \textit{"For it hath cowed my better part of man!"} (Line \#2387, Act 5, Scene 8).

\paragraph{Context.}
Macbeth's utterance immediately follows Macduff's threat to exhibit him publicly as a ``monster''
if Macbeth refuses to fight, i.e., the reaction is locally triggered by a preceding cross-speaker exchange.

\paragraph{Baseline prediction.}
The sequence baseline predicts \textit{Questioning Fate and Identity}.
While globally compatible with Macbeth's late-stage realization, it under-specifies the immediate trigger in this exchange.

\paragraph{\ce prediction and supporting structure.}
\ce ranks \textbf{\textit{Macduff as Impresario}} as the top motif,
capturing the humiliation/showman framing in Macduff's preceding threat.
In the graph, the target line is linked to (i) the local dialogue context and (ii) the speaker-interaction structure
connecting Macduff's antecedent threat to Macbeth's response.
This additional cross-speaker connectivity makes the causal antecedent topologically accessible at inference time,
favoring a fine-grained motif grounded in the exchange rather than a generic fate-related label.